\documentclass[sigconf,authorversion,nonacm]{acmart}

\AtBeginDocument{%
  \providecommand\BibTeX{{%
    \normalfont B\kern-0.5em{\scshape i\kern-0.25em b}\kern-0.8em\TeX}}}

\copyrightyear{2022}
\acmYear{2022}
\setcopyright{none}

\usepackage{subfig}
\usepackage{etoolbox}

\usepackage{algorithm}
\usepackage{algpseudocode}

\begin{document}

\title{GEO-BLEU: Similarity Measure for Geospatial Sequences}

\author{Toru Shimizu}
\affiliation{%
  \institution{Yahoo Japan Corporation}
  \city{Tokyo}
  \country{Japan}}
\email{toshimiz@yahoo-corp.jp}

\author{Kota Tsubouchi}
\affiliation{%
  \institution{Yahoo Japan Corporation}
  \city{Tokyo}
  \country{Japan}}
\email{ktsubouc@yahoo-corp.jp}

\author{Takahiro Yabe}
\affiliation{%
  \institution{Massachusetts Institute of Technology}
  \city{Cambridge}
  \state{Massachusetts}
  \country{USA}}
\email{tyabe@mit.edu}

%\vskip{1.5em}
\par
\vspace{\baselineskip}

\renewcommand{\shortauthors}{Shimizu, et al.}

\begin{abstract}
\noindent In recent geospatial research, the importance of modeling large-scale human mobility data and predicting trajectories is rising, in parallel with progress in text generation using large-scale corpora in natural language processing.
Whereas there are already plenty of feasible approaches applicable to geospatial sequence modeling itself, there seems to be room to improve with regard to evaluation, specifically about measuring the similarity between generated and reference trajectories.
In this work, we propose a novel similarity measure, GEO-BLEU, which can be especially useful in the context of geospatial sequence modeling and generation.
As the name suggests, this work is based on BLEU, one of the most popular measures used in machine translation research, while introducing spatial proximity to the idea of $n$-gram.
We compare this measure with an established baseline, dynamic time warping, applying it to actual generated geospatial sequences.
Using crowdsourced annotated data on the similarity between geospatial sequences collected from over 12,000 cases, we quantitatively and qualitatively show the proposed method's superiority.
\end{abstract}




\maketitle

\section{Introduction}
Geospatial sequence modeling over human mobility trajectories and language modeling in natural language processing (NLP) can be seen analogously, regarding places as words and human mobility trajectories as sentences.
On the geospatial side, the main workhorse is next place prediction (NPP) \cite{schreckenberger18} in which a model predicts the place a person moves to at the next time step on the basis of the past trajectory and other features, and repeating NPP while reusing predicted places as context directly leads to geospatial sequence generation.
Also, this approach can naturally extend to sequence-to-sequence or translation, assuming a model generates a trajectory using another corresponding trajectory, e.g., one in a past period, as context.
The importance of this kind of self-supervised approach is surging in geospatial research, and many modeling methods known in NLP and other related fields are feasibly applicable to geospatial problem settings.
Meanwhile, the area of evaluation still seems to be needing further consideration.

\begin{figure}
  \centering
  \includegraphics[width=0.9\linewidth]{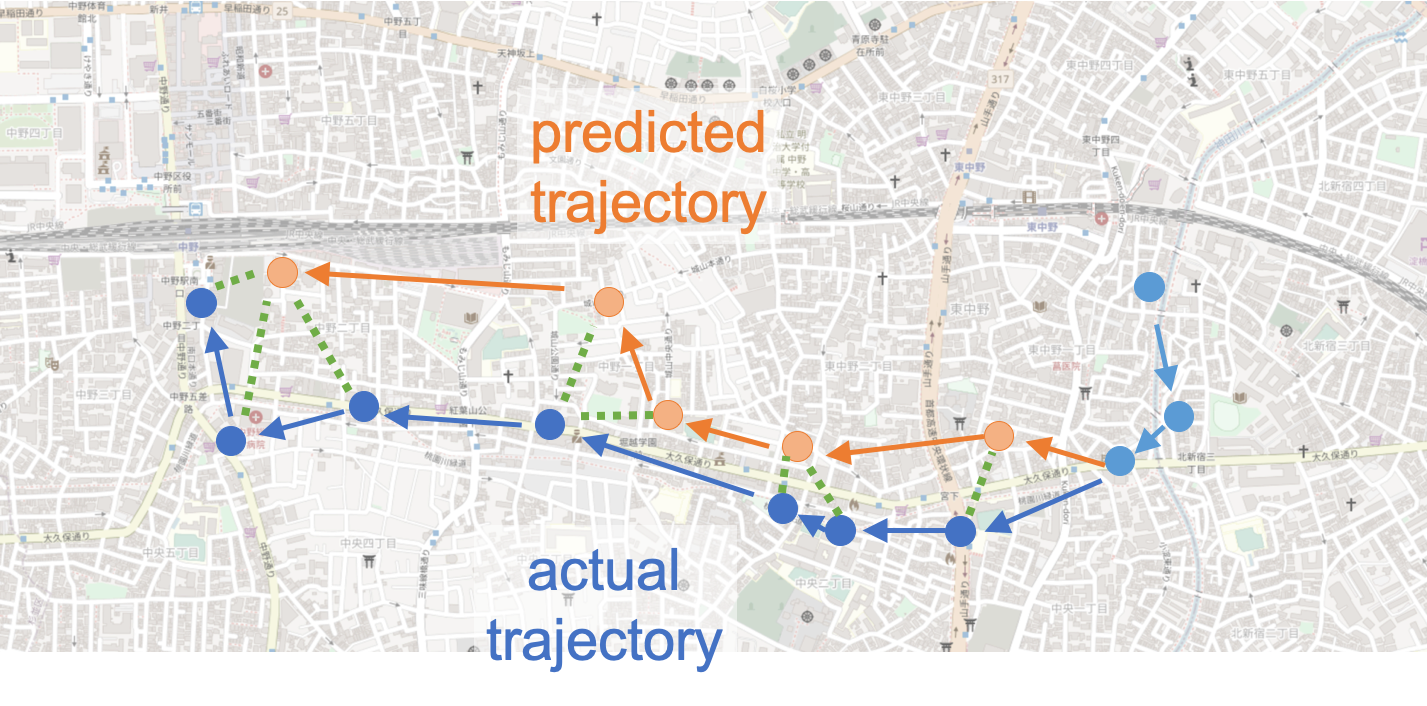}
  \caption{Predicted and actual human mobility trajectories in a relatively short time period, e.g. tens of minutes.}
  \label{fig:simple_trajectories}
\end{figure}

\begin{figure}
  \centering
  \includegraphics[width=0.9\linewidth]{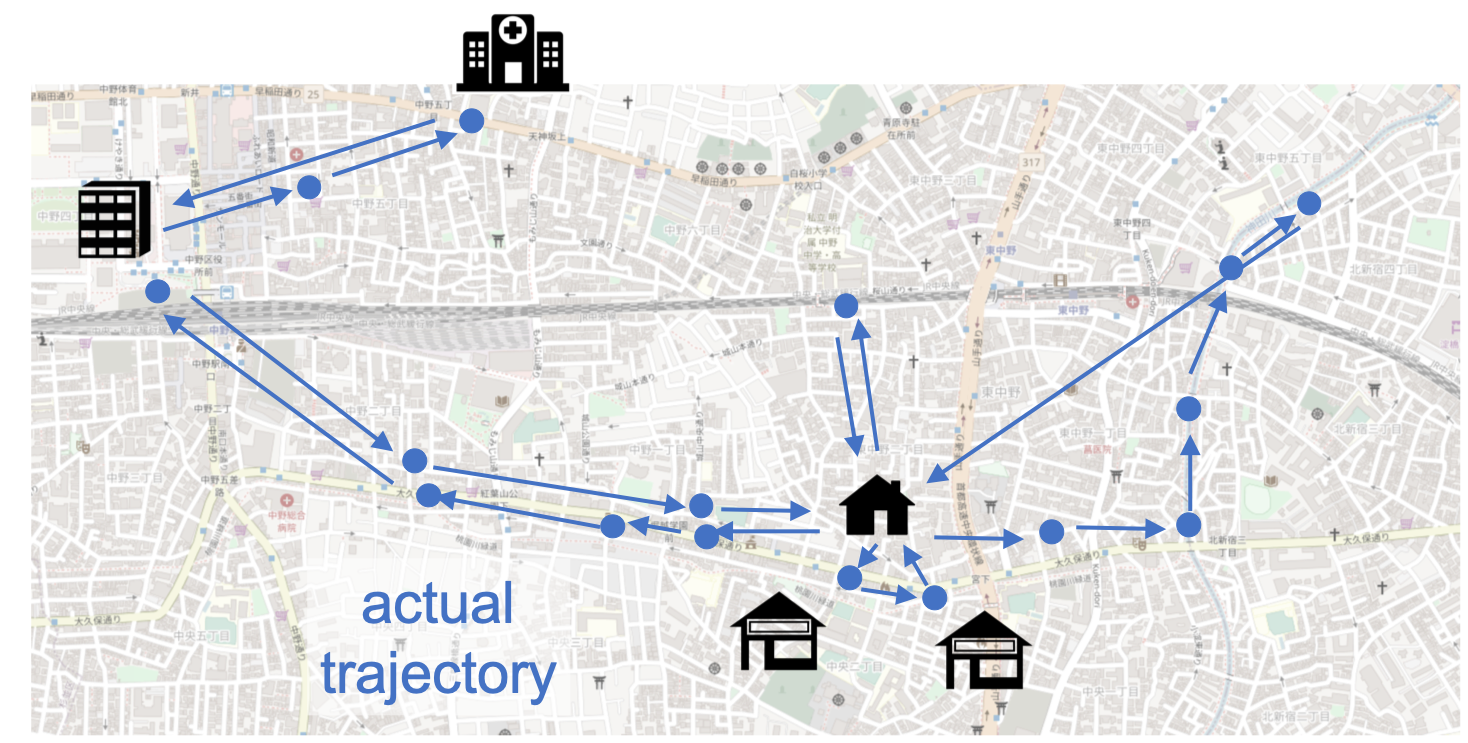}
  \caption{A complex human mobility trajectory in a relatively long time period, e.g. several days.}
  \label{fig:complex_trajectories}
\end{figure}

\begin{itemize}
  \item Dynamic time warping (DTW) \cite{vintsyuk68,sakoe78} has long been known as a way to evaluate the distance of two given sequences, and it has been used in geospatial research as well as in many other fields. 
  An essential characteristic of DTW is that it aligns the sequences for measuring entirely, without considering local features shared between them. 
  It is suitable to treat entirely aligned sequences, but not so when treating involved sequences for which step-by-step alignment does not make much sense. 
  \item BLEU \cite{papineni02} is one of the most popular measures for similarity used in NLP, especially in machine translation. 
  BLEU uses local features of given sequences, word $n$-grams, and is suitable to treat not completely aligned sequences. 
  Regarding places in sequences as words and their contiguous combinations as geospatial $n$-grams, we can apply this to evaluate the similarity of geospatial trajectories on the basis of local features. However, it has another disadvantage; the geospatial $n$-grams need to be exactly the same to be counted as ``matched'', and very close but slightly displaced ones do not contribute to the outcome.
  In other words, spatial proximity, which is potentially an important property for similarity, is not taken into account when using BLEU.
\end{itemize}

\noindent There can actually be situations where DTW is not suitable.
Figure \ref{fig:simple_trajectories} shows predicted and actual trajectories in a relatively short time period, e.g. tens of minutes.
In this case, trajectories are simple enough to be aligned in a meaningful way as illustrated by the dotted green lines, and thus DTW is applicable here without problems.
On the other hand, when the time period of prediction is relatively long, e.g. several days, trajectories to be predicted will become more involved as illustrated in Figure \ref{fig:complex_trajectories}.
The trajectory is not a straight line from one place to another anymore but a combination of subtrajectories such as one from home to the office, one from the office to a nearby hospital, and so on.
In this problem setting, we can expect that a predicted trajectory shares some motifs or subtrajectories with the actual one locally but not that the whole predicted and actual trajectories can be aligned from the start to the end in a meaningful way, possibly having subtrajectories occurring in a different order.

In this work, we propose a novel alternative, GEO-BLEU, extending BLEU to incorporate the idea of geospatial proximity into its core concept while utilizing local features and not requiring alignment.
To evaluate the measure's performance, we use a translation problem of human mobility trajectories as a plausible test case for similarity and distance measures; we collected trajectories consisting of smartphone locations % via a commercial web service provider operating in Japan 
and modeled them in a sequence-to-sequence manner.
The modeling task is to predict peoples' daily trajectories under the self-restraint of COVID-19 on the basis of trajectories in the pre-COVID-19 period, which ultimately leads to understanding behavioral implications of COVID-19 and contributing to the field of urban dynamics.
At the same time, we focus on the similarity measure in this work and stay away from stepping into this COVID-19-specific research question.
We apply our proposed measure, GEO-BLEU, and other two baselines, BLEU and DTW, to sequences generated by this translation model, comparing generated sequences and actual sequences. 
% in the mid-COVID-19 period and obtaining the scores. %もうちょい簡潔に書けそう
After that, we compare these scores with crowdsourced annotated data to quantify how consistent the measures and human intuition are, showing the proposed method's superiority.

% * 人の移動の履歴をモデリングし、seq2seqあるいは next place predictionのような形で、滞在場所のsequenceを生成するような問題設定 → 近年重要性が増している
% * 予測結果（generatedなsequence）を正解のsequenceと比較する際、評価指標の選択をどうするか
%    * sequenceを生成したモデルの与えるlossを使うことはできるが、あくまでそのモデルの「主観」によるもの
%    * DTWのような既存の手法はあるが、generatedなもの全体と正解全体の間でalignできることを前提としている → ある程度 長期間に渡るような人の行動パターンの、部分に着目した類似度の評価には適用し難い
%    * geospatialなdomain以外に目を向けると、sequenceの類似度の評価指標としては NLPにおけるBLEUが代表的。これは部分に着目した評価を行うものでもある
%       * とはいえ、言語における単語という、離散的な要素の列を扱うもの。場所のsequenceでは要素となる場所の遠近まで考慮したいが、そのような側面はBLEUそのままでは扱うことができない
% * 本研究では、BLEUにgeospatialな問題設定特有の「場所同士の遠近」を織り込んだ評価指標 GEO-BLEUを提案し、その妥当性をtoy dataおよび実際的なgenerated sequencesを用いて評価した

\section{Existing and Proposed Measures}
In this section, we first explain DTW and BLEU and then describe our proposed measure GEO-BLEU.
Also, using a toy problem, we demonstrate a notable characteristic of the proposed method.

\subsection{Existing Measures}

\subsubsection{Dynamic Time Warping.}
Dynamic time warping (DTW) \cite{vintsyuk68,sakoe78} is a distance-like measure for comparing the similarity between two sequences which was first developed in speech recognition but then has been used in various fields including geospatial research.
The method involves finding the optimal alignment between two sequences $X = (x_1, x_2, \ldots, x_M)$ and $Y= (y_1, y_2, \ldots, y_N)$.
% 以下 文中の式のフォーマット、改行まわり、投稿前要確認
One possible way of alignment is represented as a sequence of pairs between elements in $X$ and those in $Y$: $P = ((x_{i_1}, y_{j_1}), (x_{i_2}, y_{j_2}), \ldots, (x_{i_l}, y_{j_l}), \\ \ldots, (x_{i_L}, y_{j_L}))$ where $i_l \in [1:M], j_l \in [1:N]$ and $L = \max(M, N)$.
Also, there are three conditions for $P$ to be valid alignment: 
\begin{itemize}
    \item the boundary condition $(i_1, j_1) = (1, 1)$ and $(i_L, j_L) = (M, N)$, which requires the start of $X$ and $Y$ and the end of them must be matched respectively, 
    \item monotonicity condition $i_l \leq i_{l + 1}$ and $j_l \leq j_{l + 1}$ for $l \in [1: L - 1]$, which preserves the time-ordering of elements, and 
    \item step size condition $(i_{l + 1} - i_{l}, j_{l + 1} - j_{l}) \in \{ (1, 1), (1, 0), (0, 1) \}$.
\end{itemize}
The cost for such an alignment $P$ is calculated as the sum of the pairwise distance $d(x_{i_l}, y_{j_l})$:
\begin{align} \label{eq:dtw_sum}
    cost(P) = \sum\limits_{l=1}^{L} d(x_{i_l}, y_{j_l})
\end{align}
where $d(\cdot, \cdot)$ is usually the Euclidean distance between two places.
Using this, we can represent DTW as the minimum cost given by the optimal $P$:
\begin{align} \label{eq:dtw}
    DTW = \min_P cost(P).
\end{align}
As for the actual procedure of optimization, we followed a technical report \cite{senin08}.

\subsubsection{BLEU}
BLEU \cite{papineni02} is a measure being heavily used for evaluating machine translation systems for quantifying how close generated candidates are to reference human translations.
BLEU uses word $n$-grams as the unit of comparison and considers the ratio of $n$-grams matched between the generated and reference sentences to all the $n$-grams in the generated candidates for a given $n$.
The ratio, which is called modified precision $p_n$, is obtained as follows
\begin{align} \label{eq:modified_precision}
  p_n = \frac{\sum\limits_{S \in C} \; \sum\limits_{n\text{-}gram \in S} \text{Count}_{\text{matched}} (n\text{-}gram)}{\sum\limits_{S \in C} \; \sum\limits_{n\text{-}gram' \in S} \text{Count} (n\text{-}gram')}
\end{align}
where $C$ is the candidate corpus, and $S$ is each of the candidate sentences in it.
Actually, $p_n$ tends to become large when the candidates are too short.
To correct this unintended effect, BLEU uses a factor called the brevity penalty $BP$, which is given by
\begin{align} \label{eq:bp}
  BP = \begin{cases} 
    1, & \text{if $c > r$} \\
    e^{1 - r/c}, & \text{if $c \leq r$} \end{cases}
\end{align}
where $c$ is the sum of the candidates' lengths, and $r$ is that of the references.
Taking the weighted geometric average of the modified precision scores for $n \in \{1, \ldots, N\}$ while applying $BP$, resultant BLEU score $B$ is defined as
\begin{align} \label{eq:bleu}
  BLEU = BP \cdot \exp \bigg( \sum\limits_{n=1}^N w_n \log p_n \bigg)
\end{align}
where $w_n$ is the positive weight summing up to 1. 
The original work of BLEU uses $N = 4$ and $w_n = \frac{1}{N}$ for $n \in \{1, \ldots, 4\}$, and we follow the settings in the current study.
It should be noted that BLEU is for evaluating candidate and reference sentences of the whole corpus and not for evaluating a single candidate sentence.
Nevertheless, we borrow the approach of BLEU to devise a measure applicable to a single pair of sentences which can be an alternative to DTW.

% 要追記：
% * BLEUは基本 コーパス全体単位あるいはシステム単位での比較に使うもの、single sentenceには適用てきないものとされている。
% * 一方、今ほしいのは DTWの代替で single sentence (sequence)評価に使えるもの
% * BLEUをsingle sentence評価に使うことの無理は承知で、GEO-BLEUをoriginalと比べる目的で、評価に含める
% * w_nは（オリジナルの論文と同様）1固定
% * 言語系では 4-gramまで考慮することが多いが、本論文では bigramまで

\subsection{GEO-BLEU}
Our proposed measure GEO-BLEU is based on BLEU but intended to be an alternative to DTW, which means it measures a distance or similarity of a given pair of sequences.
At the same time, it borrows the concept of $n$-gram from NLP, relaxing the matching condition so that the score reflects the proximity of a given pair of $n$-grams.

As the first step, we introduce the geospatial revision of $n$-gram as a chunk of locations $(q_1, \ldots, q_n)$ where each location $q_k$ is represented as a point in two-dimensional space.
In addition, we define the similarity score $s$ of a pair of $n$-grams $g_v = (v_1, \ldots, v_n)$ and $g_w = (w_1, \ldots, w_n)$ on the basis of proximity as follows
\begin{align} \label{eq:geo-n-gram}
    s(g_v, g_w) = \prod_{k=1}^n \exp(-\beta \, d(v_k, w_k))
\end{align}
where $d(\cdot, \cdot)$ is the Euclidean distance between two locations, and $\beta$ is a coefficient for adjusting the scale.
In this manner, the similarity between $n$-grams is evaluated to become one when two $n$-grams are exactly matched. Also, the far two $n$-grams go away, the closer the value asymptotically comes to zero.
% Also, a trajectory given as a sequence of locations $(q_1, \ldots, q_L)$ can be converted into a sequence of $n$-grams $(g_1, \ldots, g_{L - n + 1})$ applying a sliding window of width $n$ to the original sequence and recognizing the $n$-grams occurring in it, as long as $L \leq n$ holds.

Next, we consider the way to match $n$-grams in the candidate sequence and those in reference.
In BLEU, the matching is conducted by the function $Count_{\text{matched}}(n\text{-}gram)$ in Equation \ref{eq:modified_precision}; it gives one if the same $n$-gram remains ``unused'' in the reference sentences, eliminating that ``used'' $n$-gram instance from the pool for subsequent matching, and otherwise gives zero.
For GEO-BLEU, which incorporates the concept of proximity, we let an $n$-gram on the candidate side form a pair with the closest unused $n$-gram remaining on the reference side, prohibiting $n$-grams on the reference side from being reused as in the BLEU's original matching rule.
We greedily optimize the set of such pairs so that the sum of the similarity scores comes close to the maximum value.
Denoting the optimized set of pairs as $P = \{(g_{c_1}, g_{r_1}), \ldots, (g_{c_L}, g_{r_L})\}$ where $L$ is the shorter of the candidate's and reference's lengths, $g_{c_k}$ is an $n$-gram of the candidate sequence, and $g_{r_k}$ is that of the reference sequence, we define our $n$-gram-based similarity $q_n$ for a pair of a candidate sequence $S$ and its reference sequence as
\begin{align} \label{eq:modified-precision-eq}
    q_n = \frac{\sum\limits_{(g_c, g_r) \in P} s(g_c, g_r)}{\sum\limits_{n\text{-}gram \in S} Count(n\text{-}gram)}
\end{align}
which can be computed as shown in Algorithm \ref{arg:q_n}.
Taking the weighted geometric mean for a range of $n$ in the same manner as Equation \ref{eq:bleu} and introducing the brevity penalty $BP$ as in Equation \ref{eq:bp}, the proposed similarity measure GEO-BLEU is given as
\begin{align} \label{eq:geo-bleu}
    GEO\text{-}BLEU = BP \cdot \exp \bigg( \sum\limits_{n=1}^N w_n \log q_n \bigg).
\end{align}
In our experiments, we use $\beta = 1$, $N = 3$, and $w_n = \frac{1}{N}$ for $n \in \{1, 2, 3\}$.

If BLEU is applied to evaluating a single candidate, there can be cases in which the modified precision becomes zero.
On the contrary, the modified-precision equivalent of GEO-BLEU always has a non-zero value due to the relaxed matching, and this property makes GEO-BLEU more feasible and suitable for evaluating a single candidate sequence.

\begin{algorithm}[tb]
\caption{Computation of Equation \ref{eq:modified-precision-eq} including greedy matching of $n$-grams in candidate and reference sequences}
\label{arg:q_n}
\begin{algorithmic}[1]

\State Input: two arrays of $n$-grams, candidate $C = \{{g_c}_1, \ldots, {g_c}_{L_c}\}$ and reference $R = \{{g_r}_1, \ldots, {g_r}_{L_r}\}$
\State $similaritySum \gets 0$
\State Create array $A$, initially empty
\ForAll{$g_c, g_r \in C \times R$}
    \State $similarity \gets s(g_c, g_r)$ using Equation \ref{eq:geo-n-gram}
    \State Append tuple $(g_c, g_r, similarity)$ to $A$
\EndFor
    
\State Sort $A$ by $similarity$ in decreasing order
\While {$A$ is not empty}
    \State $g_c, g_r, similarity \gets A[0]$
    \State $similaritySum \gets similaritySum + similarity$
    \State Remove all the tuples involving $g_c$ or $g_r$ from $A$ \footnotemark
\EndWhile
\State \Return $similaritySum / L_c$

\end{algorithmic}
\end{algorithm}
\footnotetext{The $n$-grams in $C$ and $R$ are distinct identifiers representing positions of the sliding window of width $n$ moving over the original sequences of places. Therefore, in this step, a tuple $({g_c}', {g_r}', \cdot)$ is not removed from $A$ just because $g_c$ and ${g_c}'$ (or $g_r$ and $g_r'$) are referring to the same actual chunk of places. It is removed only when the identifiers themselves are matched.}

% GEO-BLEUの取りうる最小値、最大値について補足

% ---

% * p_nを出した後の式はBLEUと同じ
% * n-gramのmatchのカウント方式を変える
%   → 例の expで減衰するやつ、式書いておく
%   → candidate側とreference側のgeo-n-gramがペアになっているとする。geo-n-gramにはそれぞれn個の要素があるが、対応するもの同士の距離に着目。n個の距離を「一致する場合に1、離れると減衰」な、expな関数に入れて 積をとったものそのペアのスコアとする
%   → geo-n-gramペア単位で考えても、完全一致で 1、離れると漸近的に0に近づく、という良い性質に

% * マッチングについて： さらに、BLEUの「各n-gramのmatch数カウントは、そのn-gramのcandidate側の出現数を超えない」という性質を「candidate側のn-gramとreference側のn-gramの 近いもの同士でgreedyにマッチ」というアルゴリズムで再現

% より詳しく述べると： BLEUにおけるcandidateとreferenceの n-gramマッチは「candidate側の各n-gramが reference側の相方を探し、見つかればペア形成（success）、見つからなかったりあぶれたりすれば（reference側n-gramは再利用できず）fail」という動作とみなすことができる。match数とはすなわちsuccess数。
% 同様のことを、GEO-BLEUでは「candidate側の各n-gramがreference側の相方、近いものを探し、見つかればペア形成（success）、遠いものでもペアは形成できるが あぶれるケースは出てきてfailに該当」という動作と定義。match数に相当するのは、ここでは「ペア全体での、geo-n-gramのマッチ評価値合計」。
% ペア形成は全体最適なアプローチも考えられるが、計算時間も考慮して「ペア候補全体から最もgeo-n-gramのマッチスコアが高いもの優先」で一つずつgreedyに行う
% 最初に（順序を考慮しない）全ペア候補を列挙しておき、それぞれについてスコア計算。ペア形成されたものがあれば、candidate側 reference側 それぞれのgeo-n-gramが参加しているペアを候補から削除
% これを candidateあるいはreferenceの残存geo-n-gramがなくなるまで繰り返す
% p_nの分母はBLEUと変わらず、candidate側のgeo-n-gram数合計

\subsubsection{Characteristics of GEO-BLEU}

\begin{figure}
  \centering
  \includegraphics[width=0.7\linewidth]{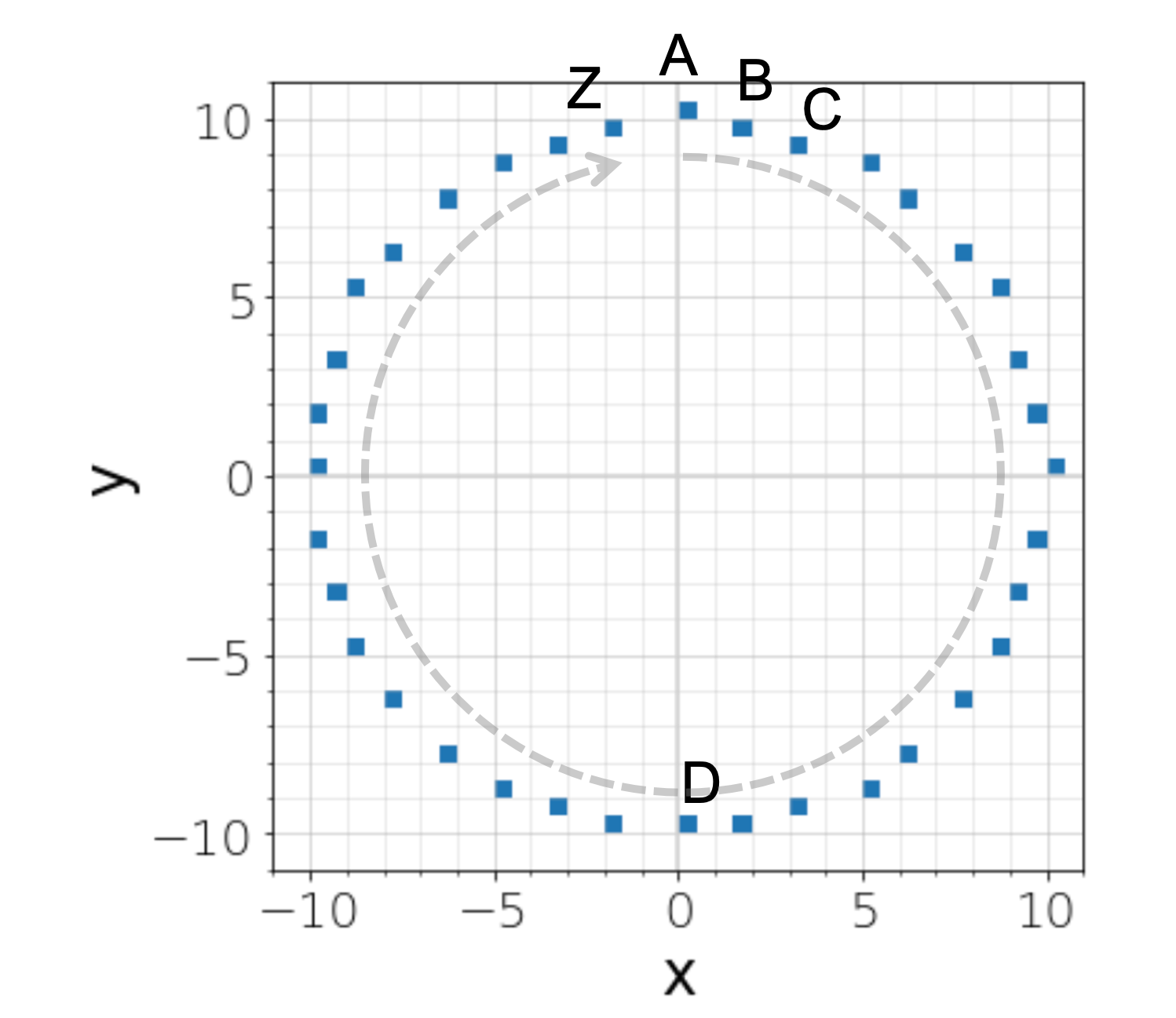}
  \caption{A sample sequence consisting of 36 grid cells placed over a circle of 10 km radius.}
  \label{fig:toy_sequence}
\end{figure}

\begin{figure}
  \centering
  \includegraphics[width=0.8\linewidth]{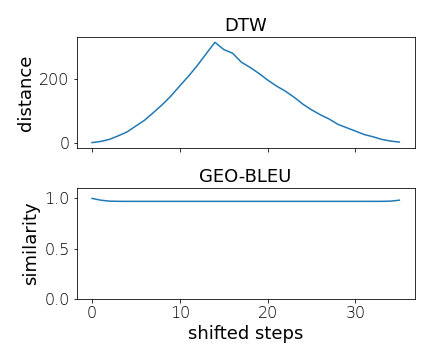}
  \caption{The scores of DTW and GEO-BLEU depending on the extent of the phase shift between the original and its shifted, derived sequence.}
  \label{fig:toy_seq_plot}
\end{figure}

To illustrate the characteristics of GEO-BLEU and its difference from DTW, we apply the two measures to simple toy sequences in two-dimensional space and compare the results.
As shown in Figure \ref{fig:toy_sequence}, we consider 36 grid cells with sides of 0.5 km placed over a circle of 10 km radius at almost regular intervals.
Our original sample sequence starts from cell A, goes clockwise through B, C, and the following, and ends at Z as shown as the dashed arc arrow.
Then, by moving the start and end points clockwise and one step at a time, i.e., by shifting the phase forward, we can generate variations such as one going clockwise from B to A, another from C to B, and so on for evaluating the similarity with or distance from the original.
Here, it is crucial that whether they are similar or different depends on the evaluations' purpose and point of view, and there is no definite criterion in that regard.
Considering the original sequence and another with the opposite phase starting from D, they are completely different when aligned wholly.
In this view, the distance between the first cells of the sequences is 20 km, the maximum possible number in this setting, and it does not change in the following aligned pairs, such as one between the second cells of the two sequences.
On the other hand, those two sequences can be seen as almost identical when concerned with the local features, as they share almost all the cells and chunks except for those around the start and end.
Among these conflicting points of view, GEO-BLEU is for comparing sequences on the basis of local features as in the latter example, while DTW views two sequences wholly aligned as in the former.

Figure \ref{fig:toy_seq_plot} shows the actual distance calculated by DTW and similarity by GEO-BLEU between the original and shifted sequences where the $x$-axis denotes the number of the shifted steps.
The subject of comparison is the original sequence itself at $x = 0$, two sequences have the opposite phases at around $x = 18$, and the phase difference becomes very small again at the rightmost point $x = 35$. %comment-in
The results are contrasting; the value of DTW is significantly changing depending on $x$,
% the number of shifted steps 
while that of GEO-BLEU is staying around the maximum possible value as two sequences are always similar considering their local features.
As illustrated, GEO-BLEU is a measure for comparing sequences on the basis of their partial or local features and without aligning them wholly.

% 円のやつ、phaseがずれるとDTWは急減、対して GEO-BLEUはノーダメージ

% さらに、「とあるシーケンスについて、減衰関数を極限までsteepにしてみる」数値実験を行う？ BLEUとGEO-BLEUで値が同じになることを示せるはず。多分。

\section{Experiments}
%%% covidの影響を翻訳モデルで捉え、前期間のsequenceから 後期間のそれを生成、正解と比較、、という一連の流れについて軽く述べておく

\subsection{Human Mobility Trajectory Dataset}
For this study, a web service company, Yahoo! JAPAN, provided us with smartphone GPS records of their users, which had originally been collected for their services.
The users have agreed to provide their location information for research purposes, and the data are anonymized so that individuals cannot be identified and that personal properties such as gender and age are unknown. 
Each GPS record consists of a user's ID, timestamp, longitude, and latitude.

% 以下2文、comment-in
One thing to note here is that the trajectory data is inherently noisy to some extent from various reasons such as the error of GPS sensors and timing for sending the location data which is adjusted by the user's situation and smartphone's conditions.
Thus, it is not guaranteed that two identical actual trajectories turn into the identical sequences in the obtained data.

Using this smartphone GPS data, we extracted records from two consecutive periods, one from Oct. 1st, 2019 to Mar. 31st, 2020 and the other from Apr. 1st to May 25th in 2020.
The periods were determined so that the data captures two different modes of the society in Japan, one mode without the influence of COVID-19, which corresponds to the former period, and the other mode under nationwide self-restraint of activities to prevent the spread of the infection, which corresponds to the latter period.
As for deciding the specific dates, we referred to dates of the government's relevant announcements in addition to public stringency index data \cite{ritchie20}.

From this set of GPS records, we prepared one million pairs of trajectories representing how the mobility pattern of a person in the pre-COVID-19 period has changed in the mid-COVID-19 period.
In this data preparation process, the longitudes and latitudes in the GPS records are aggregated and discretized into 500m-square grid cells on an hourly basis so that a sequence of grid cells corresponds to a trajectory of a person.
Using this set of GPS records, we obtained 1 million pairs of sequences as follows:
\begin{itemize}
  \item aggregated the longitudes and latitudes so that it represents a trajectory of the person,
  keep trajectories spreading over both periods and discard the rest
  \item summarized and discretized the longitudes and latitudes into 500m-square grid cells representing hourly positions, and
  \item took out the longest partial sequence which is contained within the Tokyo metropolitan area for each of two periods of a pair so that the sequence does not contain unknown places.
\end{itemize}
Then, we allocated 10,000 pairs to the validation set, another 10,000 pairs to the test set, and the rest to the training set.
The resultant vocabulary size, i.e. the number of unique 500m-square cells within the area appearing in the sequences, is 10,584.
Also, The average length of sequences is 90.5 in the former period and 124.5 in the latter period.
Considering that each step stands for an hourly position, each of these sequences usually amounts to a mobility pattern spanning over several days rather than a short trajectory within a day.

% * smartphoneの位置情報から生成した、人の移動のsequence
% * 日本の東京近郊における、「COVIDの影響を受ける前の、人の移動のsequence」を、「COVIDで自粛している状態の、人の移動のsequence」に翻訳
%    * 前期間： 2019/10/1〜2020/3/31
%    * 後期間： 2020/4/1〜2020/5/25
%    * 期間を考える際の参考情報： 
% * 場所をgridとして捉える、メッシュサイズ 500m
%    * gridのcell一つ一つにidが振られており、それが sequenceの各要素として、入力および予測の対象として用いられる
% * sequenceの作り方： 最大長、UNKの有無（無し）、前期間・後期間からsequenceを抜き出す際の考え方（ランダムなところから取る、必ずしも 前期間の最後・後期間の最後 ではない）など
% * データセットのサイズ、train / validation / test
% * vocabulary size
\begin{figure*}
  \centering
  \includegraphics[width=.75\linewidth]{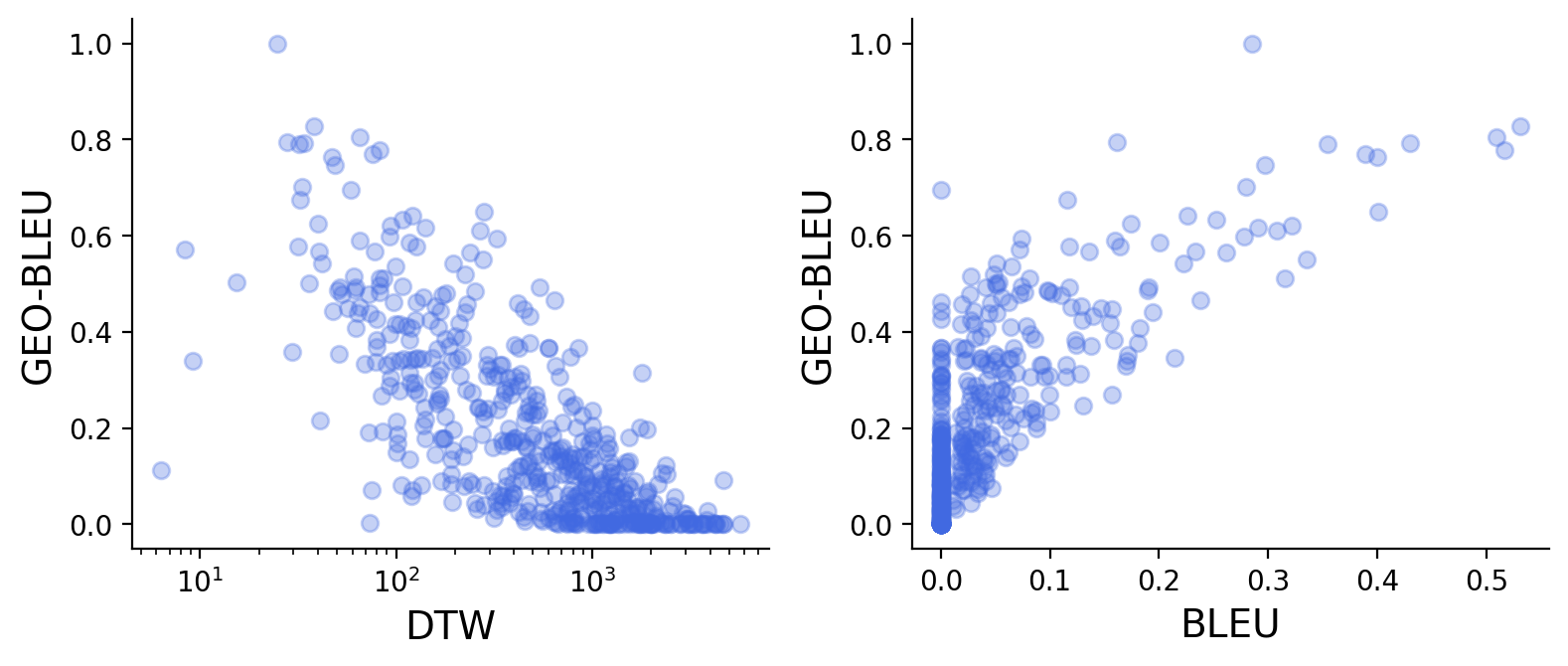}
  \caption{Comparison of GEO-BLEU scores with DTW and BLEU scores, respectively, for the 500 pairs of generated and actual human mobility trajectories.}
  \label{fig:comparison}
\end{figure*}

\begin{figure*}
  \centering
  \includegraphics[width=.85\linewidth]{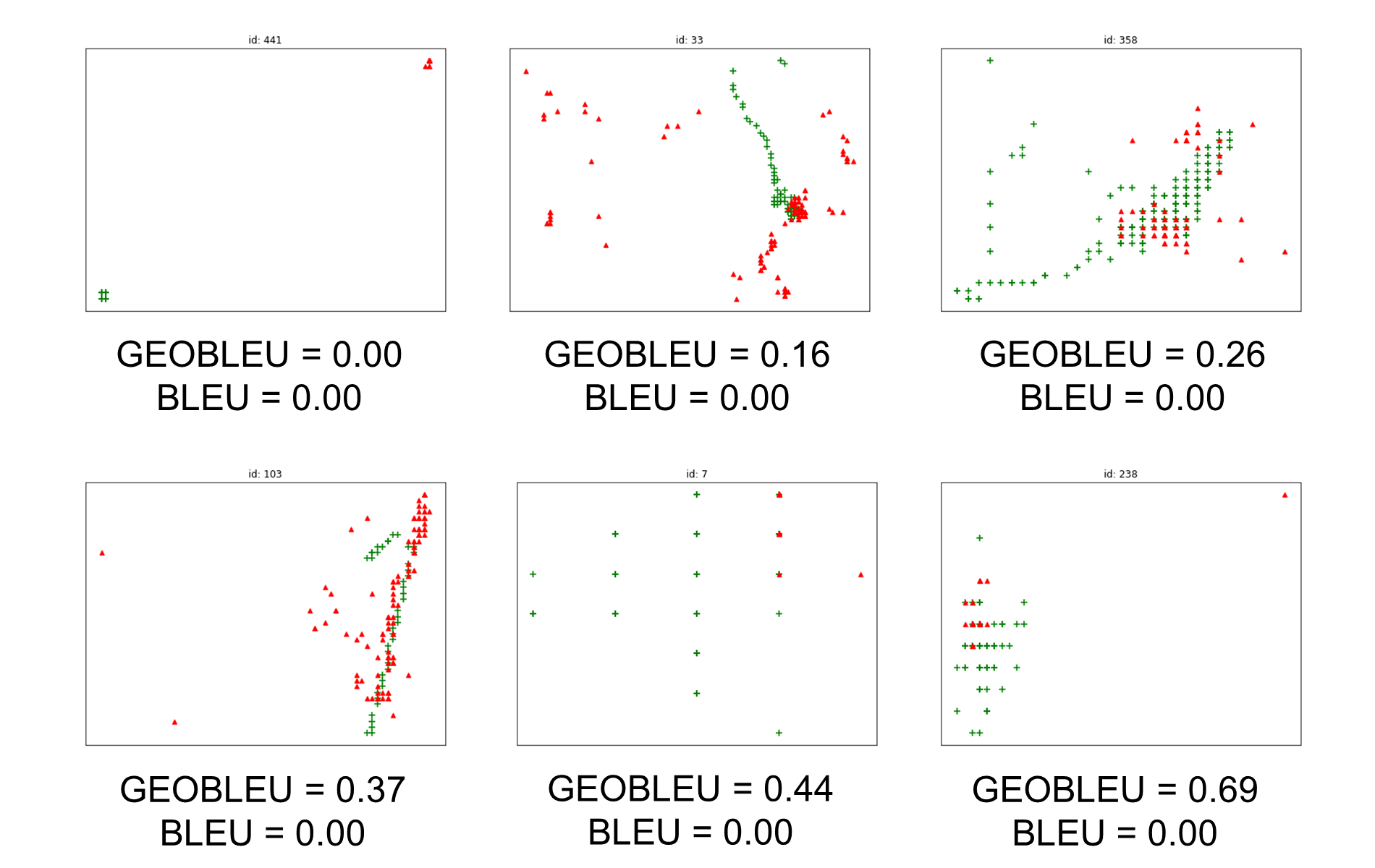}
  \caption{Examples of trajectory pairs where the BLEU scores are 0, but GEO-BLEU scores are non-zero (except for the top left one). GEO-BLEU is capable of capturing subtle differences and similarities compared to BLEU.}
  \label{fig:bleuzero}
\end{figure*}

\subsection{Model and Training}
We trained a seq2seq model \cite{sutskever14,cho14} consisting of a pair of two-layer LSTM-RNNs using the dataset for ten epochs so that it can generate a person's trajectory in the mid-COVID-19 period given a trajectory in the pre-COVID-19 period.
After the training finished, we took out a model with the lowest validation loss and applied it to sample sequence generation as in the next section.
% We consider that this approach comes under self-supervised learning as the preparation of the prediction targets does not involve any human judgment or annotations, whereas the model is seq2seq, which is often applied to supervised translation problems.

% * LSTM RNN
% * 前期間のsequenceを読み込み、分散表現を生成、それをスタート地点として next place predictionを行う形で後期間のsequenceを予測。seq2seqあるいは encoder-decoder architectureの範疇
% * sequenceの各要素において、場所以外の時間などの要素も入力（しているが今回 本質ではない、、）
% * 予測対象は場所のみ
% * 入力sequenceの長さはいろいろ、出力は 200ステップ固定

%%% この比較についても少し語る。BLUEは０付近に多いけどGEOBLEUはそういうケースの微小な違いとかも拾えてる。

\subsection{Evaluation and Results}

\begin{figure*}
  \centering
  \includegraphics[width=.6\linewidth]{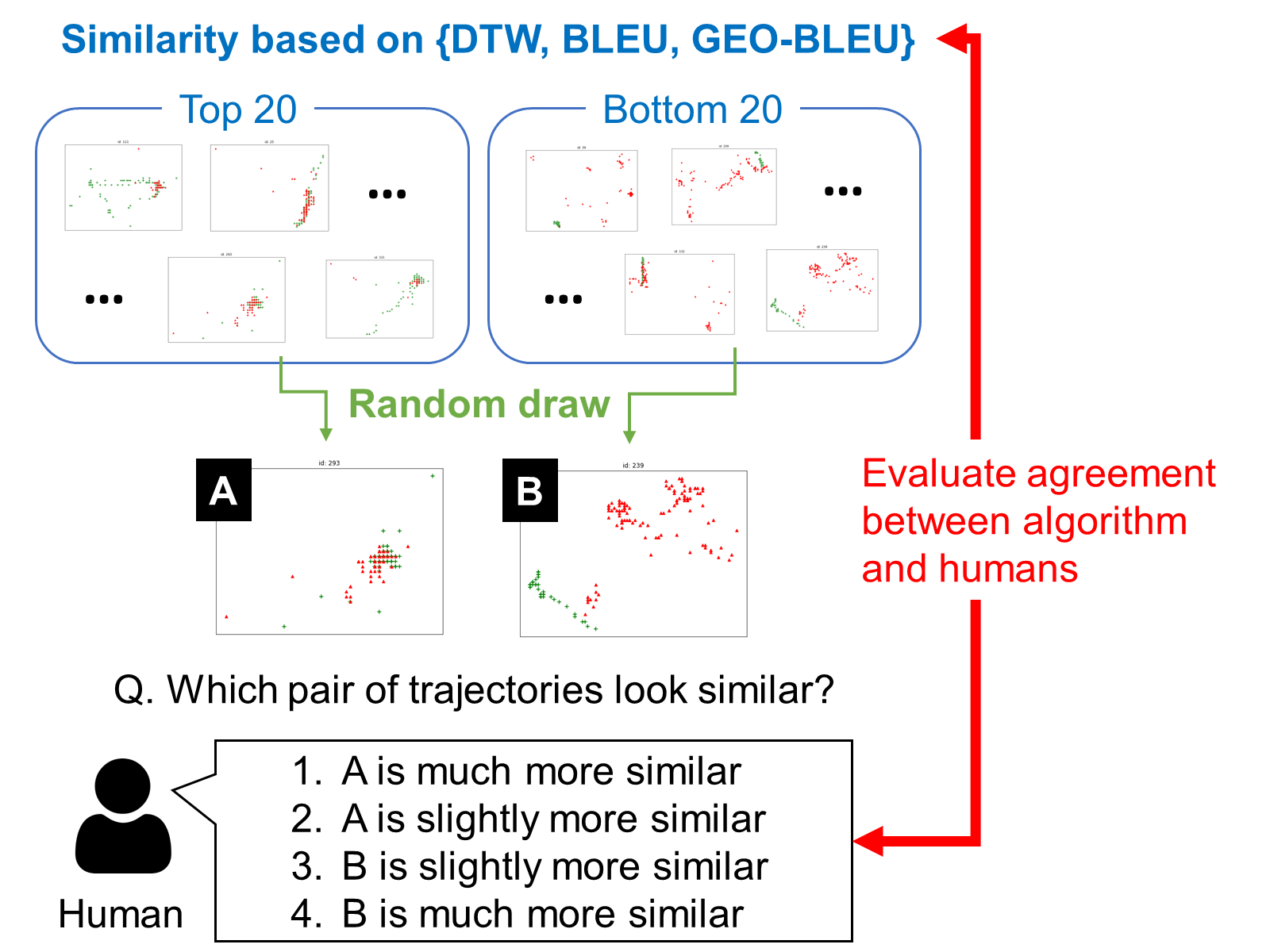}
  \caption{Outline of crowdsourcing framework used to validate the performance of GEO-BLEU.}
  \label{fig:test}
\end{figure*}

\begin{figure}
  \centering
  \includegraphics[width=0.95\linewidth]{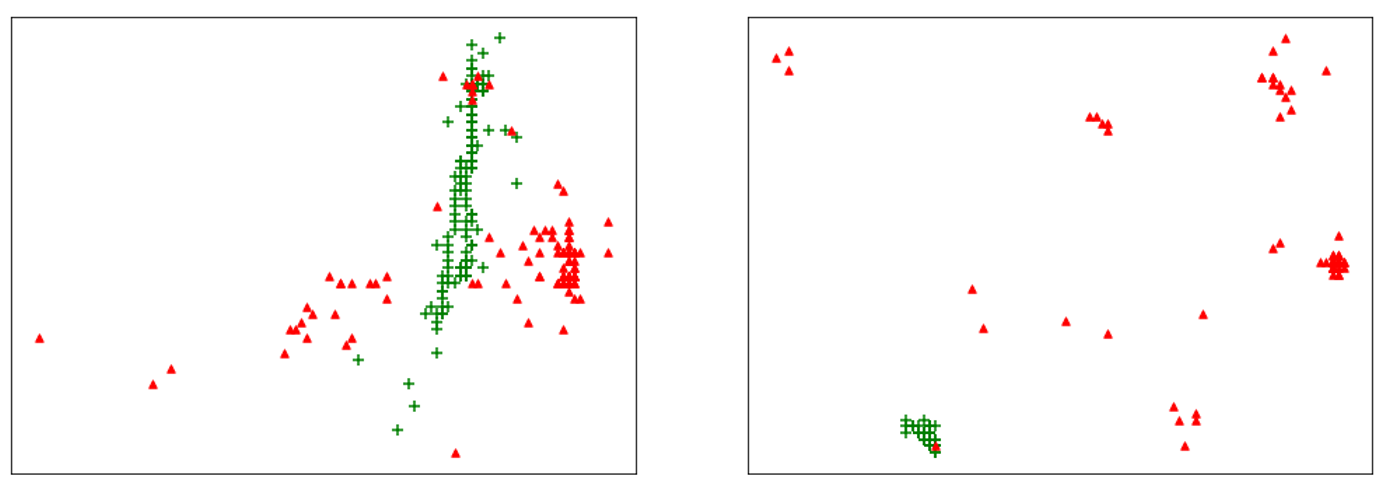}
  \caption{An example of annotation cases for evaluating each measure. The red dots show the steps of a generated trajectory, and the green dots show the steps of its corresponding actual trajectory. In this example, the left side actually belongs to the top-20 pairs and the right side the bottom-20 pairs, and the annotator is asked to distinguish which side comes from the top-20. The easier the cases of a measure are, the better the measure is considered to be.}
  \label{fig:annotation}
\end{figure}

\begin{table}
  \centering
    \caption{The average of annotation scores for each measure.}
  \begin{tabular}{lccc}
  \toprule
  Method & normalized DTW  & BLEU & \bf{GEO-BLEU}\\ \midrule
  Score & 0.550 & 0.530 & \bf{0.699}  \\ \bottomrule
  \end{tabular}
  \label{tbl:eval_results}
\end{table}

\begin{figure*}[!h]
  \centering
  \includegraphics[width=.75\linewidth]{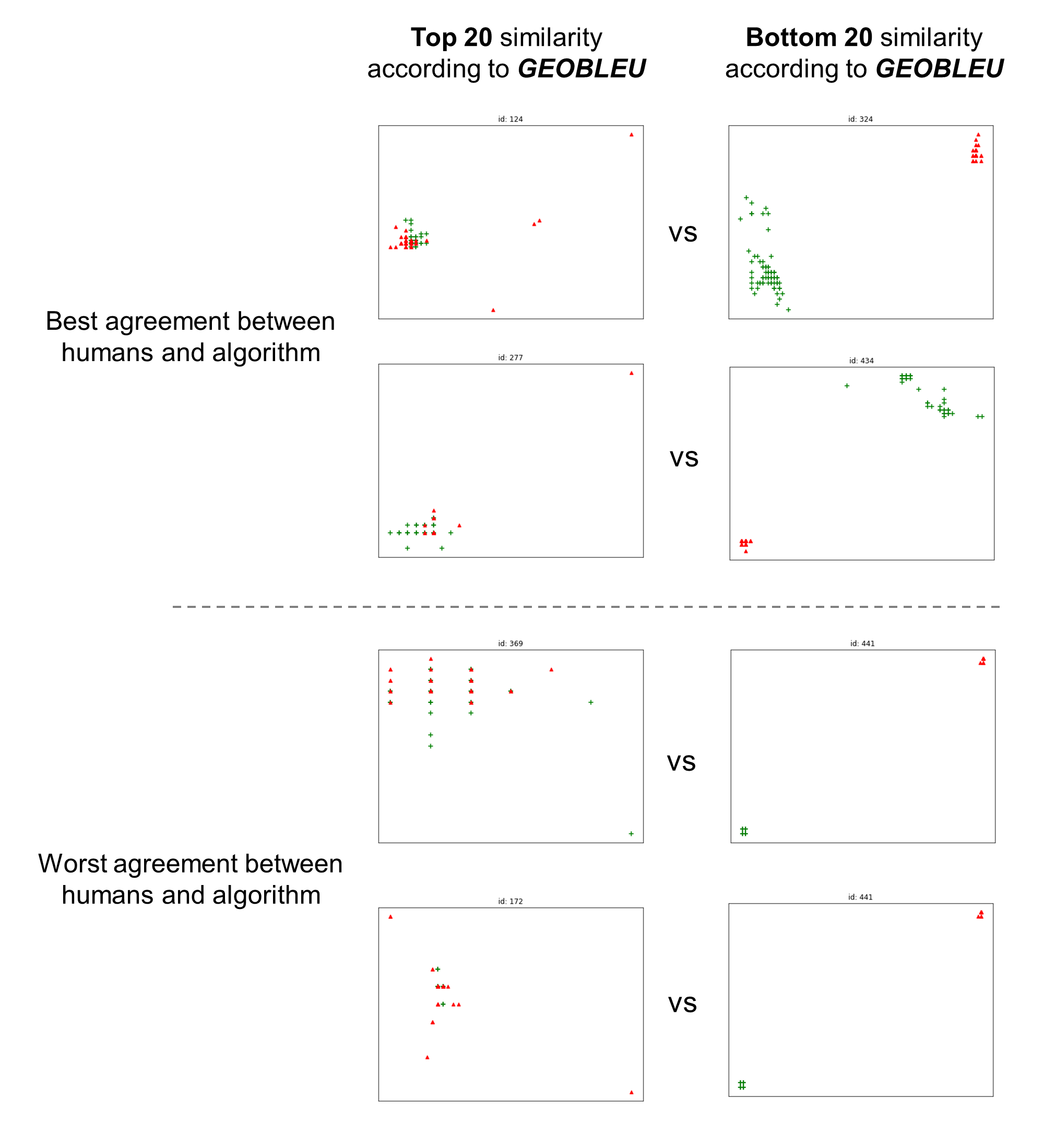} 
  \caption{Comparison of cases with the most frequent agreements and disagreements between the humans and GEO-BLEU scores.}
  \label{fig:geobleuerrors}
\end{figure*}

\begin{figure*}[!h]
  \centering
  \includegraphics[width=.75\linewidth]{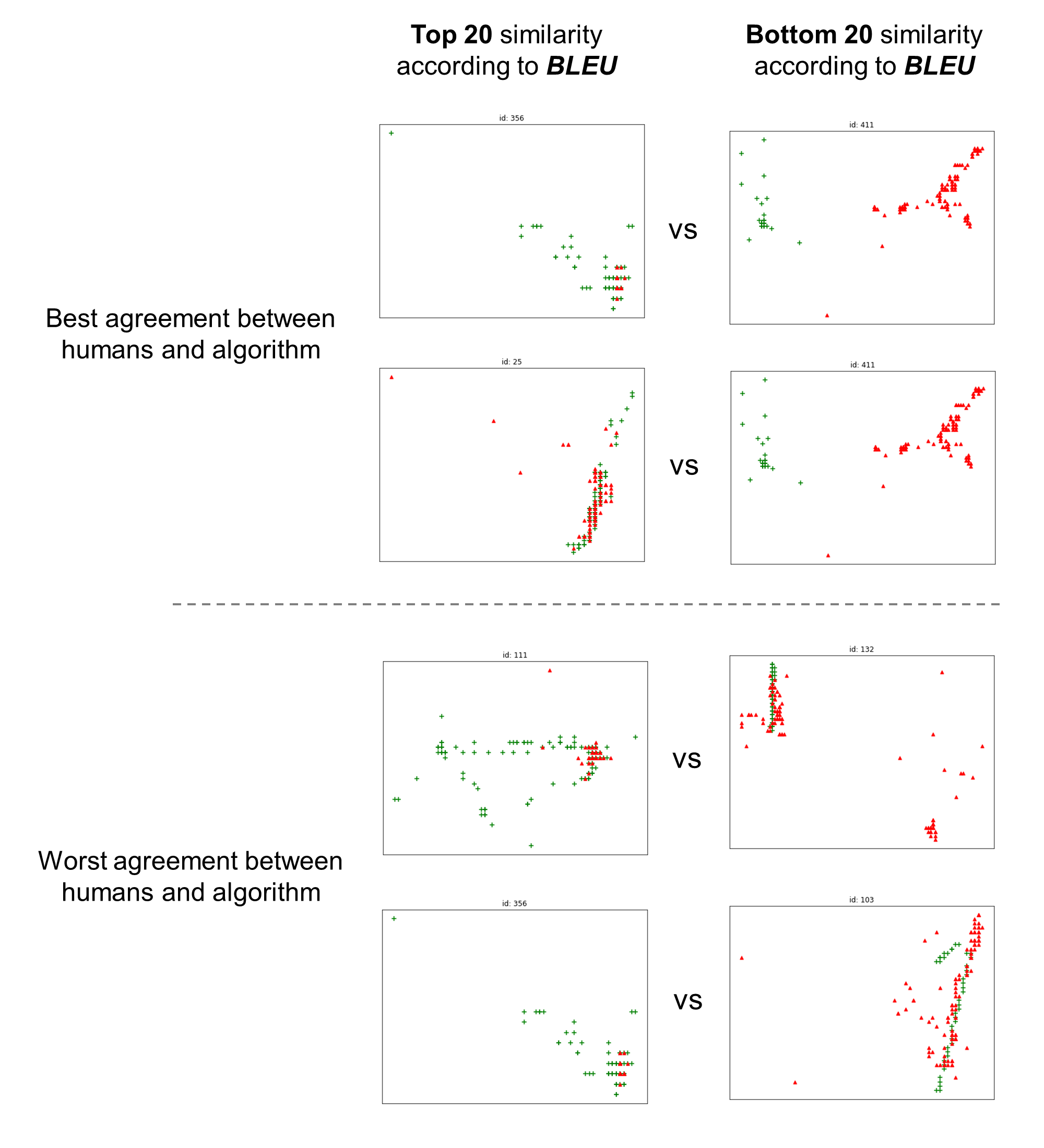} 
  \caption{Comparison of cases with the most frequent agreements and disagreements between the humans and scores.}
  \label{fig:bleuerrors}
\end{figure*}

We generated 500 sample sequences using the trained model and the test set.
Then, we scored each sequence with three measures, normalized DTW, GEO-BLEU, and BLEU, comparing the generated trajectory with the actual one in the dataset.
As DTW has a dependency on the sequence length as in Equation \ref{eq:dtw_sum}, we normalize the raw scores dividing them by $L$.
Actually, BLEU is a quality measure not of a single sentence but of the entire corpus.
However, we dared to apply BLEU for evaluating each sequence here for reference, treating a pair of sequences as a small corpus, to show the difference between the original BLEU and GEO-BLEU.
% It should be noted that GEO-BLEU having a better result than BLEU in this evaluation does not mean that the former is superior to the latter.
Figure \ref{fig:comparison} shows the DTW, GEO-BLEU, and BLEU scores for each of the samples in the dataset. We observe a negative correlation between DTW and GEO-BLEU and a positive correlation between BLEU and GEO-BLEU, which is intuitive. 
Comparing BLEU to GEO-BLEU, we see that GEO-BLEU is able to assign positive scores to instances where the BLEU score is zero, since GEO-BLEU is designed to capture the subtle differences (where BLEU assigns a zero score even for small distance differences). 
Figure \ref{fig:bleuzero} shows some examples where BLEU gave a score of zero, however GEO-BLEU gave some positive scores, ranging from 0 to 0.69. We can see here how GEO-BLEU gives partial credit for having some overlap or closeness in the predicted data over the ground truth data.

\subsection{Crowdsourced Experiment}

To evaluate how convincing the scores given by a measure are, we sorted the 500 pairs of generated and actual trajectories by each of three measures into descending order of similarity, obtaining three lists of the same entries but in different orders.
Then, we took out the top-20 pairs and the bottom-20 pairs from those lists.
There are $20 \times 20 = 400$ possible combinations between the top pairs and bottom pairs for each measure, and for each such combination, we asked annotators which pair of generated and actual sequences look more similar, showing the top and bottom pairs side-by-side, % while randomly shuffling the sides, 
as shown in Figure \ref{fig:annotation}.
An annotation is given as one of four options: the left is clearly similar, the left is somewhat similar, the right is somewhat similar, and the right is clearly similar.
We assign a positive score to a case if the judgment is consistent with the measure: 1.0 if the top-side is judged as clearly similar and 0.5 if it is somewhat similar.
If the judgment is inconsistent with the measure, the score becomes -1.0 for ``clearly similar'' and -0.5 for ``somewhat similar'' to give a penalty.
We presented one case to ten different annotators and collected $400 \times 10 \times 3 = 12,000$ judgments. The overall procedure of the crowdsourced experiment is shown in Figure \ref{fig:test}.
Table \ref{tbl:eval_results} shows the averaged score for three measures, and GEO-BLEU is superior to normalized DTW in this comparison.
While the purpose of BLEU is not a single-sequence measurement as in this evaluation, the scores themselves imply that GEO-BLEU derived from it is modified so that it becomes more suitable to the current problem settings. 

Figures \ref{fig:geobleuerrors} and \ref{fig:bleuerrors} show examples where the labels given by humans and the algorithms (GEO-BLEU and BLEU, respectively) had the largest and least agreement. The top two rows of Figure \ref{fig:geobleuerrors} show two example pairs where both the algorithm (GEO-BLEU) and humans indicated that the trajectories on the left are much more similar than the one on the right. The lower two rows of Figure \ref{fig:geobleuerrors} are examples where humans indicated that the trajectory pairs on the right hand side were more similar compared to the ones on the left, which is not intuitive. We suspect that human labelers mistook the meaning of `similarity of two trajectories' in these cases.

% * この後のsequence生成・評価では、（overfitする手前の）bestなモデルと、そこに至るまでの生煮えモデルを用いて比較
% * モデル毎の normalized DTW、BLEU、GEO-BLEU
% * visualizeした結果
% * （計算時間について書く？）
% * trg_sys sequence長は 125固定、実際のデータの平均長と同じ

\section{Related Work}
Many studies have proposed methods to measure the similarity of two movement trajectories. First, there are two major types of cases: one is to compare and evaluate the entire movement trajectory, and the other is to compare and evaluate a part of the movement trajectory. The former is called complete match measure and the latter is called partial match measure, as summarized in the following section.
Other types of measures have already been proposed as described in a survey \cite{su2020survey}.
Still, to the best of our knowledge, this work is the first to apply the concept of ``geospatial $n$-gram'' to such evaluation, to take into account the local features of sequences.

\subsection{Complete Match Measure}
Complete match measure is a method of comparing two movement trajectories with respect to all measurement points. For example, if the two movement trajectories are perfectly matched, the distance between them is zero. However, in many cases, there are differences between the two movement trajectories, and the amount of those differences is compared. The difficulty with this method of comparing movement trajectories is that it must take into account not only distance but also time differences. In addition, the length of the movement trajectory changes on a case-by-case basis. For instance, if trajectory A is 10 km in 2 hours and trajectory B is 130 km in 8 hours, it is difficult to simply compare the two trajectories.

The most basic method for a complete match is the Euclidean distance \cite{ed2-1,ed2-2}, which calculates the difference in the norm of the trajectories to be compared.
It was first proposed as a distance measure between time series and was once considered as one of the most widely used distance functions since the 1960s \cite{ed1-1,ed1-2}.
It is now also used to evaluate movement trajectories. In this case, the trajectories must have the same length.

The most famous algorithm for complete match is Dynamic Time Warping (DTW) \cite{bhadane2017analysis,cai2018trajectory}. This method has been used for a long time to measure distances in time series data \cite{dtw1, dtw2,dtw3, dtw4}, and it is now also used to compare movement trajectories. The algorithm is simple \cite{dtw0}, and the lengths of the two trajectories do not have to be the same.

\subsection{Partial Match Measure}
Partial match measure is a method to measure similarity in only one part of two movement trajectories with a large amount of information.
Two well-known methods for partial match are the Longest common subsequence (LCSS) and edit distance on real sequence (EDR) methods.

First, LCSS measures \cite{lcss1,lcss2} the length of the sequence common to two trajectories at successive points. For example, two people who were separated at the start meet at a certain point, travel the same distance for a while, and then break up again. In this case, the LCSS method does not consider the degree of separation between the two trajectories, but focuses only on the common trajectory, and the longer the trajectory, the more similar the trajectories are.

EDR \cite{edr1,edr2} is a method to calculate how much processing of the movement trajectory A should be done to match the movement trajectory B. For example, the similarity is defined as the cost of repeatedly performing insertions, deletions, or substitutions until the two match. The greater the processing cost, the lower the similarity between the two movement trajectories. Many proposals have been made regarding the definition of processing methods and costs.

Methods such as LCSS and EDR, which measure similarity only for a part of the movement trajectory, are often effective as distance measures without being complicated to process. However, the goal of this research is to measure similarity as perceived by humans. People often make judgments of similarity by looking at the entire image, and partial match is not appropriate for this purpose.

% distance metrics \cite{larios2015evaluating}. \\

% BLEU \cite{papineni02}. \\
% DTW \cite{vintsyuk68,sakoe78}. \\
% LSTM \cite{hochreiter97}. \\

% \section{Discussion}

% - how to ask humans % アンケートの設問による勘違い防止

% - 

\section{Conclusion}
We proposed a novel similarity measure of sequences, GEO-BLEU, extending BLEU by incorporating proximity into the core concept and using place $n$-grams as local features so that it can evaluate the similarity of predicted and reference trajectories without aligning them step by step from the start to the end.

The effectiveness and characteristics of GEO-BLEU was tested on two types of data sets: a set of artificial data and an in-the-wild data set based on actual user movement trajectories.
In a realistic setting about self-supervised geospatial sequence modeling, GEO-BLEU is more consistent with annotators' intuition for similarity than an existing popular measure, DTW.
The proposed GEO-BLEU should be applicable to many future studies in diverse research fields as a practical evaluation index for similarity of spatial trajectories in general.

%%% 汎用性について追記、DTWが広い範囲で使われているのと同様、これも・・・
%%% さらに、corpus単位・モデル単位での評価にも使える旨、あらためて（手前のどこかでも書いておくとよさそう）
%%% 今回、BPは使ってないので、このままの形でよいか future workにて要評価

\bibliographystyle{ACM-Reference-Format}
\bibliography{geobleu}

\end{document}